# Fine-Grained Vehicle Classification in Urban Traffic Scenes using Deep Learning

Syeda Aneeba Najeeb[1], Rana Hammad Raza[1], Adeel Yusuf[1] and Zamra Sultan[1]

[1] Department. of Electronics and Power Engineering
Pakistan Navy Engineering College
National University of Sciences and Technology, Karachi, Pakistan

`(syeda.najeeb2015, hammad, adeel)@pnec.nust.edu.pk,`
`zsultan.beee15pnec@student.nust.edu.pk`

**Abstract.** The increasingly dense traffic is becoming a challenge in our local settings, urging the need for a better traffic monitoring and management system. Fine-grained vehicle classification appears to be a challenging task as compared to vehicle coarse classification. Exploring a robust approach for vehicle detection and classification into fine-grained categories is therefore essentially required. Existing Vehicle Make and Model Recognition (VMMR) systems have been developed on synchronized and controlled traffic conditions. Need for robust VMMR in complex, urban, heterogeneous, and unsynchronized traffic conditions still remain an open research area. In this paper, vehicle detection and fine-grained classification are addressed using deep learning. To perform fine-grained classification with related complexities, local dataset THS-10 having high intra-class and low interclass variation is exclusively prepared. The dataset consists of 4250 vehicle images of 10 vehicle models, i.e., Honda City, Honda Civic, Suzuki Alto, Suzuki Bolan, Suzuki Cultus, Suzuki Mehran, Suzuki Ravi, Suzuki Swift, Suzuki Wagon R and Toyota Corolla. This dataset is available online. Due to having almost no design variation in some make and models over the years, vehicle models are not separated by their year of generation. Two approaches have been explored and analyzed for classification of vehicles i.e, fine-tuning, and feature extraction from deep neural networks. A comparative study is performed, and it is demonstrated that simpler approaches can produce good results in local environment to deal with complex issues such as dense occlusion and lane departures. Hence reducing computational load and time, e.g. finetuning Inception-v3 produced highest accuracy of 97.4% with lowest misclassification rate of 2.08%. Finetuning MobileNet-v2 and ResNet-18 produced 96.8% and 95.7% accuracies, respectively. Extracting features from fc6 layer of AlexNet produces an accuracy of 93.5% with a misclassification rate of 6.5%.

# 1    Introduction

Image-based Vehicle Make and Model Recognition (VMMR) systems keep their importance in computer vision as they provide the foundation of a smart traffic monitoring system. This includes vehicles' color, year of production, model and vehicle make classification. Fine-grained vehicle classification appears to be a challenging task as compared to vehicle coarse classification. High intra-class variance and low inter-class variance are two major challenges in this area. This task becomes further challenging in local environment due to the lack of a local dataset, and heterogonous, irregular and unsynchronized vehicle movement.

Automatic make and model recognition from frontal images of cars has been reported for fine-grained classification by authors at [8]. The research community has proposed the use of HOG, SIFT, and SPM for vehicle make recognition [9]-[11]. Use of 3D model has also been tested for this purpose [12]. Medioni and Prokaj presented the approach for feature comparison using 3-D models, while 3D curve alignment is used by Ramnath et al. [13]. In recent years, recognition of vehicle make and model using deep learning has been efficiently demonstrated by Lee et al. [14]. Yu et al. used Faster R-CNN [15] for fine-grained vehicle classification while combination of different CNNs has been used by Ma et al. [16] for the same purpose.

In this paper, two approaches have been demonstrated and analyzed for the classification of vehicles i-e by fine-tuning of CNNs and feature extraction from different layers of CNNs. A local dataset acquisition task was exclusively conducted for training and testing purposes. The dataset is named as THS-10 (i.e 10 models of Toyota, Honda and Corolla).

# 2    Dataset

For the collection of data, video captured by a surveillance camera installed on a local bridge is acquired. Three videos have been recorded, each on separate days, with varying traffic and lighting conditions. The road has five lanes in total, having extremely dense traffic with each vehicle moving at medium speed. Vehicles in the scene have frontal and slightly side views as shown in Figure 1. Summary of one of these videos is given in Table 1. Acquired video feed is passed to YOLO v2 for vehicle detection. These vehicles are then cropped out from each detected output. Dataset is generated by manually separating vehicle images into ten vehicle categories. It consists of a total of 4250 images of 10 vehicle classes. The dataset is publicly available on 'https://github.com/ZamSul/Fined-Grained-Vehicle-Classification-in-Urban-Traffic-Scenario-using-Deep-Learning' and its details are provided in Table 2.

# 3    Methodology

The detected vehicle images are first pre-processed to make valid input size. Later they are fed as input into network's first layer for training. Once training is completed, the network's performance is evaluated on test dataset which predicts output class with class probability. Experimental results are obtained using two approaches i.e fine-tuning deep neural networks and extracting useful features from different layers of CNNs for training them on SVM.

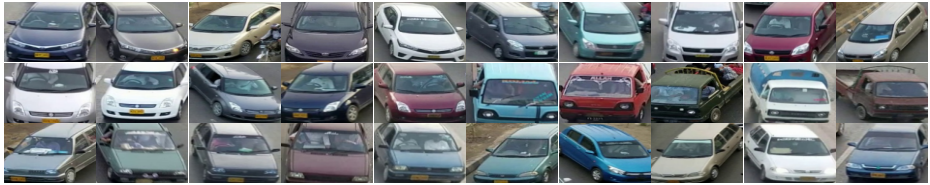
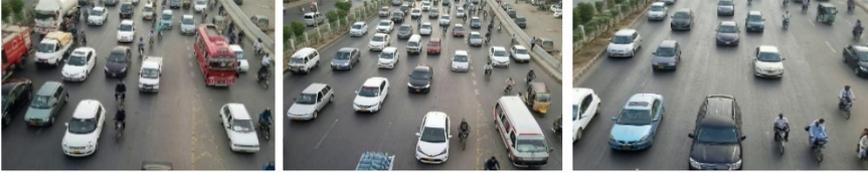

| Sequence Type | Outdoor | Camera View | Top frontal and slightly side angle |
|---|---|---|---|
| Video Length | 01:00:00 | Camera Motion | Static |
| Environment | Sunny | Frames per Second | 30 |
| Object Class | Vehicle | Shadow Size | Small |
| Object Size | Medium | Shadow Strength | Low |
| Object Speed | Medium to Fast | Noise Level | Low |

**Table 2.** Summary of THS-10 Dataset

| Vehicle Class | Total Extracted Images | Training Images | Testing Images |
|---|---|---|---|
| Honda City | 485 | 390 | 95 |
| Honda Civic | 404 | 325 | 79 |
| Suzuki Alto | 382 | 327 | 55 |
| Suzuki Bolan | 450 | 388 | 62 |
| Suzuki Cultus | 486 | 391 | 95 |
| Suzuki Mehran | 460 | 382 | 78 |
| Suzuki Ravi | 397 | 319 | 78 |
| Suzuki Swift | 390 | 331 | 59 |
| Suzuki Wagon R | 371 | 311 | 60 |
| Toyota Corolla | 425 | 358 | 67 |

In the first approach, results are obtained by fine-tuning neural networks that are already trained on ImageNet dataset consisting of 1000 classes. These neural networks are then trained and tested on THS-10 dataset, to find the best fit. Visualizing and examining each layer's activations for the second approach provides a better understanding about network learning. For latter approach, features are extracted from different layers of neural networks (such as AlexNet, GoogleNet, ResNet, and Inception-v2), which are then trained on SVM classifier to get classification results. Results for both approaches are evaluated and presented in Section 4.

## 4     Results and Analysis

In this section, the performances of deep neural networks have been evaluated using accuracy and misclassification. Accuracy is defined as the ratio of correctly classified instances whereas misclassification rate is the ratio of incorrectly classified instances during the course of classification. They are calculated as

$$\text{Accuracy} = \frac{\text{True Positive} + \text{True Negative}}{\text{Total Instances}}, \quad \text{Misclassification Rate} = \frac{\text{False Positive} + \text{False Negative}}{\text{Total Instances}}$$

Seven different neural network models i.e. AlexNet[1], SqueezeNet[5], ShuffleNet[6], GoogleNet[2], ResNet-18[3], MobileNet-v2[7] and Inception-v3[4] are fine-tuned on local THS-10 dataset for classifying vehicles images into ten vehicle classes. Inception-v3 produced highest accuracy of 97.4 % and a misclassification rate of only 2.03%. Summary of all results is provided in Table 3.

**Table 3.** Classification accuracy by fine-tuning deep neural networks on THS-10 Dataset

| Deep Neural Network (Fine-Tuned) | # of Network Layers | Classification Accuracy (%) | Misclassification Rate (%) |
|---|---|---|---|
| AlexNet | 8 | 90.6 | 9.4 |
| SqueezeNet | 18 | 91.9 | 8.1 |
| ShuffleNet | 50 | 92.2 | 7.8 |
| GoogleNet | 22 | 93.5 | 6.5 |
| ResNet-18 | 18 | 95.7 | 4.3 |
| MobileNet-v2 | 53 | 96.8 | 3.2 |
| **Inception-v3** | **48** | **97.4** | **2.03** |

Features are extracted from different layers of deep neural networks with performances evaluated on THS-10 dataset. In ResNet-18, low-level shallow features extracted from initial layers have produced poor accuracy. Most effective and rich features are found to be of Fc6 layer of AlexNet with accuracy of 93.5% and layer 822 of Inception-v2-ResNet with a classification accuracy of 93.4%. From different experiments, it is observed that initial layers of network learn very general representation of features like edges, etc. Higher layers contain rich features of input dataset. Fc6 and fc7 layers of AlexNet, Pool5-7x7_s1 layer of GoogleNet, Pool_5, Res3b_Relu and fc1000 of ResNet-18, Pool 5 layer of ResNet-101 and average pooling layer of Inception-v2 have been explored for classification. Summary of results is provided in Table 4.

*Misclassification Analysis*: Misclassification appears due to either low inter-class variance or high intra-class variance. From confusion matrix, misclassification is observed between Honda City, Honda Civic, and Toyota Corolla. This is due to a very low inter-class variance between these classes. As apparent from Figure 2 (a, b & c), their shape and frontal structure resemble each other to a great extent. Similarly, Suzuki

Ravi and Suzuki Bolan contribute to misclassification rate due to low inter-class variance. Misclassification due to high intra-class variance is observed in Suzuki Alto, as shown in Figure 2(d). Both images in Figure 2(d) show Suzuki Alto, but they are different in shape and size due to differences in models. Suzuki Wagon R, Suzuki Swift, Suzuki Mehran, and Suzuki Cultus, on the other hand, have very low intra-class variance, and therefore low misclassification rates are observed for these vehicle classes.

**Table 4.** Classification accuracy achieved extracting useful features from deep neural networks on THS-10 Dataset

| Deep Neural Network | Feature Extraction Layer | Classification Accuracy (%) | Misclassification Rate (%) |
|---|---|---|---|
| AlexNet | fc7 | 92.9 | 7.1 |
| **AlexNet** | **fc6** | **93.5** | **6.5** |
| GoogleNet | Pool5-7x7_s1 | 93.1 | 6.9 |
| ResNet-18 | Pool_5 | 89.4 | 10.6 |
| ResNet-18 | Res3b_Relu | 70.2 | 29.8 |
| ResNet-18 | fc1000 | 86.1 | 13.9 |
| Inception-v2-ResNet | Avg pool | 93.4 | 6.6 |
| ResNet-101 | Pool 5 | 92.4 | 7.6 |

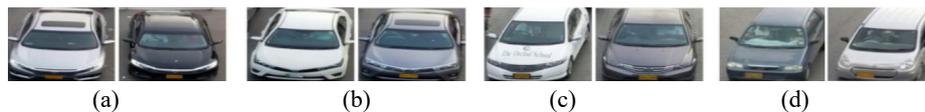

(a)          (b)          (c)          (d)

**Fig. 2.** Sample Images of classes with low inter-class variance. (a) Honda Civic (b) Toyota Corolla (c) Honda City, Sample Images for classes with high intra-class variance (d) Suzuki Alto

## 5 Conclusion

In this paper, performance of two classification schemes has been evaluated on locally generated dataset. First one is by fine-tuning deep neural networks, and second is by extracting high-level features from different layers of deep neural networks and classifying them by SVM. According to results achieved, fine-tuning of CNN outperformed the accuracies achieved by feature extraction. After implementation of proposed approach, some misclassifications were observed due to high intra-class variance or low inter-class variance. Intra-class variance can be minimized, and inter-class distance can be maximized simultaneously by using a deeper network for classification. Adding occluded instances in dataset can further reduce the dense occlusion issue. As a future work, algorithms like super-resolution can be integrated to tackle surveillance-related low-resolution problems.

## 6 Acknowledgement

We acknowledge partial support from National Center of Big Data and Cloud Computing (NCBC) and Higher Education Commission (HEC) of Pakistan for conducting this research.

# 7 References

1. Krizhevsky A., Sutskever I, Hinton GE. "ImageNet classification with deep convolutional neural networks." Advances in Neural Information Processing Systems. vol.25 (2012): 1097-1105 .
2. Szegedy C., Liu W., Jia Y., Sermanet P., Reed S., Anguelov D., Erhan D., Vanhoucke V. and Rabinovich A. "Going deeper with convolutions." IEEE Conference on Computer Vision and Pattern Recognition (CVPR), Jun. 2015, pp. 1-9.
3. He K., Zhang X., Ren S. and Sun J., "Deep residual learning for image recognition." IEEE Conference on Computer Vision and Pattern Recognition (CVPR), Jun. 2016, pp. 770-778.
4. Xie S., Sun C., Huang J., Tu Z. and Murphy K.,"Rethinking the inception architecture for Computer Vision." IEEE Conference on Computer Vision and Pattern Recognition (CVPR), Jun. 2016, pp. 2818-2826.
5. Iandola F.N., Han S., Moskewicz M.W., Ashraf K., Dally W.J. and Keutzer K., "SqueezeNet: AlexNet-level accuracy with 50x fewer parameters and< 0.5 MB model size" arXiv preprint arXiv:1602.07360 (2016).
6. Zhang, X., Zhou, X., Lin, M. and Sun, J., "ShuffleNet: An extremely efficient convolutional neural network for mobile devices." IEEE Conference on Computer Vision and Pattern Recognition (CVPR) 2018.
7. Howard A.G., Zhu M., Chen B., Kalenichenko D., Wang W., Weyand T., Andreetto M. and Adam H., "MobileNets: Efficient convolutional neural networks for mobile vision applications" arXiv preprint arXiv:1704.04861 (2017).
8. Pearce, G. and Pears, N., "Automatic make and model recognition from frontal images of cars." 2011 8th IEEE International Conference on Advanced Video and Signal-based Surveillance (AVSS) IEEE, Sep. 2011, pp. 373-378.
9. Ma, X. and Grimson, W.E.L., "Edge-based rich representation for vehicle classification," 10th IEEE International Conference on Computer Vision (ICCV), vol. 2. Oct. 2005, pp. 1185–1192.
10. Peng, Y., Yan, Y., Zhu, W. and Zhao, J., "Vehicle classification using sparse coding and spatial pyramid matching," IEEE 17$^{th}$ International Conference on Intelligent Transportation Systems (ITSC), Oct. 2014, pp. 259–263.
11. Llorca, D.F., Arroyo, R. and Sotelo, M.A., "Vehicle logo recognition in traffic images using HOG features and SVM," IEEE 16th International Conference on Intelligent Transportation Systems (ITSC), Oct. 2013, pp. 2229–2234.
12. Khan, S.M., Cheng, H., Matthies, D. and Sawhney, H. "3-D model-based vehicle classification in aerial imagery," IEEE Conference on Computer Vision and Pattern Recognition (CVPR), Jun. 2010, pp. 1681–1687.
13. Ramnath, K., Sinha, S.N., Szeliski, R. and Hsiao, E., "Car make and model recognition using 3D curve alignment," IEEE Winter Conference on Applied Computer Vision. (WACV), Mar. 2014, pp. 285–292.
14. Lee, H.J., Ullah, I., Wan, W., Gao, Y. and Fang, Z. "Real-time vehicle make and model recognition with the residual SqueezeNet architecture." Sensors, vol.19.5 (2019): 982.
15. Yu, S., Wu, Y., Li, W., Song, Z. and Zeng, W. "A model for fine-grained vehicle classification based on deep learning." Neurocomputing, vol.257 (2017): 97-103.
16. Ma, Z., Chang, D., Xie, J., Ding, Y., Wen, S., Li, X., Si, Z. and Guo, J., "Fine-grained vehicle classification with channel max pooling modified CNNs." IEEE Transactions on Vehicular Technology, vol.68.4 (2019): 3224-3233.